\documentclass{article}

    \PassOptionsToPackage{numbers, compress}{natbib}

\usepackage[ruled,vlined]{algorithm2e}
\usepackage[final]{neurips_2021}
\usepackage{graphicx}

\usepackage[utf8]{inputenc} 
\usepackage[T1]{fontenc}    
\usepackage{url}            
\usepackage{booktabs}       
\usepackage{amsfonts}       
\usepackage{nicefrac}       
\usepackage{microtype}      
\usepackage{xcolor}         

\usepackage{caption}
\usepackage{amsmath}
\usepackage{amssymb}

\title{Post-discovery Analysis of Anomalous Subsets}

\author{%
  Isaiah Onando Mulang' \\
   IBM Research Africa\\
   \texttt{mulang.onando@ibm.com} \\
   \And
   William Ogallo\\
   IBM Research Africa \\
   \texttt{william.ogallo@ibm.com } \\
   \And
   Girmaw Abebe Tadesse \\
   IBM Research Africa \\
   \texttt{girmaw.abebe.tadesse@ibm.com } \\
   
   \And
   Aisha Walcott-Bryant \\
   IBM Research Africa \\
   \texttt{alwalcott@ke.ibm.com} \\
}

\begin{document}

\maketitle

\begin{abstract} 
Analyzing the behavior of a population in response to disease and interventions is critical to unearth variability in healthcare as well as understand sub-populations that require specialized attention, but also to assist in designing future interventions. Two aspects become very essential in such analysis namely: i) Discovery of differentiating patterns exhibited by sub-populations, and ii) Characterization of the identified subpopulations.  For the discovery phase, an array of approaches in the anomalous pattern detection literature have been employed to reveal differentiating patterns, especially to identify anomalous subgroups. However, these techniques are limited to describing the anomalous subgroups and offer little in form of insightful characterization, thereby limiting interpretability and understanding of these data-driven techniques in clinical practices. In this work, we propose an analysis of differentiated output (rather than discovery) and quantify anomalousness similarly to the counter-factual setting. To this end we design an approach to perform post-discovery analysis of anomalous subsets, in which we initially identify the most important features on the anomalousness of the subsets, then by perturbation, the approach seeks to identify the least number of changes necessary to lose anomalousness.   Our approach is presented and the evaluation results on the 2019 MarketScan Commercial Claims and Medicare data, show that extra insights can be obtained by extrapolated examination of the identified subgroups.

\end{abstract}

\section{Introduction} 


Healthcare is known for its variations in the population, treatment, and outcome domains ~\cite{appleby2011variations,krumholz2013variations}. Examples of such variations include diversity in the population (e.g., a particular subgroup being vulnerable or at high risk), heterogeneous effects of a treatment regime across a given population, or different interventions across population segments in response to a specific disease. 

To improve clinical practice and the quality of interventions, there is a need to trace and reconstruct such variations. However, large-scale observational health data and longitudinal clinical records are characterized by complex interactions of patient characteristics and interventions, making challenging to analyze and interpret non-random variations of patient populations, care delivery, and outcomes. Manual approaches heavily rely on domain expertise that requires pre-assumption of features responsible for variations and lack scalability for large and complex interactions expected in clinical data.  In addition, manual approaches cannot readily unearth complex, non-obvious variations in healthcare. 

Fortunately, researchers have sought to develop automatic and data-driven approaches for the scalable discovery of differentiated patterns of care. 
Particularly, machine learning techniques in the anomalous pattern detection literature could be employed to encode variations in healthcare. 
Subset scanning ~\cite{neill2012fast,neill2013fast} is one of these approaches that aim to identify anomalous subgroups in a population with divergent characteristics compared to the expected baseline in an efficient manner. In subset scanning, anomalousness is quantified by a scoring function, typically a log-likelihood ratio statistic, that is maximized over the subsets in a multidimensional array to identify the subset with the highest score ~\cite{neill2012fast}. This scoring function must satisfy the Linear-Time Subset Scanning (LTSS) so that the search is efficiently conducted in linear time~\cite{neill2012fast}. 

The typical input to a subset scanning algorithm is a set of discrete/discretized features e.g. \textit{\textbf{Gender}: Male/Female, \textbf{Race}: Black/Brown/White, \textbf{Smoking}: Yes/No, and \textbf{Weight}: Low/Mid/High}; and target variable of interest e.g. \textit{Mortality}.
The typical output of subset scanning consists of 3 main items: 1) the highest scoring (most anomalous) subset characterized by specific feature values (e.g. \textit{\textbf{Gender}: Female, \textbf{Race}: Black/White, \textbf{Smoking}: Yes, and \textbf{Weight}: High}), (2) the score of the highest-scoring subset, and (3) the empirical p-value evaluating the significance of the identified subset. This output is thus merely descriptive 
and there is still a significant concern related to the characterization of the findings into practices, particularly in clinical care. The characterization includes understanding the dynamics associated with anomalousness of a particular sub-population in response to variation in patient features.  In other words, there is an implicit next step necessary to assist better utilization of these methods and inform actions.

In this paper, we propose the \textit{post-discovery} analysis of anomalous subsets aimed at enhancing the clinical usefulness of subset scanning results and triggering a paradigm shift from description to actionable insights. 
To this end, we tackle two major research questions as our contributions:
\begin{enumerate}
    \item Which of the features (and their values) describing the identified subset have the greatest impact on the anomalous score?
    \item What is the smallest change in feature values that causes a loss in anomalousness?
\end{enumerate}

This paper describes our initial step towards a larger agenda of post-discovery analysis of anomalous subsets. Our proposed solution involves a two-step approach 
that first identifies the features and feature values that contribute most significantly to the anomalous behavior of the identified subset, then performs cross-substitution of these feature values to attain a significant drop in anomalousness. The rest of this paper is structured as follows: Section \ref{related-work} offers a summary of related work followed by the details of the proposed method in Section~\ref{sec:methods}. The data and experiments are described in Section \ref{experiment}, finally, Section \ref{conclusion} concludes and offers future directions.

\section{Related Work}
\label{related-work}
Although our work is an initial step towards post-discovery analysis, being the first proposal in this direction, it lies in the intersection between two fields namely: Subgroup analysis and counterfactual explanations. We view our work as a special form of counterfactual analysis that operates on unsupervised scenarios. Hereafter, we discuss existing work in these two areas in literature.

\subsection{Subset Scanning}
Subset Scanning algorithms~\cite{neill2012fast,somanchi2017,mcfowland2013fast} identify the subset of records with the most significant difference in the distribution of the observed outcome compared to the expected outcome and have been applied across a variety of use cases. Neil et al. applied subset scanning in disease surveillance to detect geolocations and data streams suggestive of emerging outbreaks ~\cite{neill2012fast, neill2013fast}. Zheng et al. developed a subset scanning approach for detecting predictive bias in binary classifiers ~\cite{zhang2016identifying}. McFowland et al. proposed a subset scanning algorithm for detecting heterogeneous treatment effects in clinical trial data ~\cite{mcfowland2018efficient}. Somanchi et al. developed the anomalous patterns of care scan that combines propensity score modeling with subset scanning to enable anomalous pattern detection in anomalous pattern detection in observational health data~\cite{somanchi2017}. Each of these methods optimizes a scoring function, such as a log-likelihood ratio statistic, over all subsets in a dataset and selects the highest scoring subset. The scoring function is constrained to a  mathematical property referred to as Linear-Time Subset Scanning (LTSS) that enables the search to be conducted in linear as opposed to exponential time ~\cite{neill2012fast}.

\subsection{Counterfactual Analysis}
Traditionally, counterfactual explanations have been employed to help understand the limitations of machine learning methods. 
To this end, methods for counterfactual explanations try to find an example that is as close as possible to the data points in the training set of a model. 
Counterfactual explanation methods can either be model-specific or model-agnostic.
Model-specific methods utilize the internal training characteristic (e.g., the hyperparameters) of the underlying machine learning model ~\cite{DBLP:journals/corr/abs-1907-02584,10.1145/3351095.3372850, 10.1145/3287560.3287569}. Model-agnostic methods separate the explanations from the machine learning model. They encompass functionality that apply to any arbitrary models\cite{DBLP:journals/corr/abs-1906-00117, Karimi2020ModelAgnosticCE,10.1145/3375627.3375812,White2020MeasurableCL}, generally by only accessing the prediction function of a fitted model ~\cite{Dandl2020MultiObjectiveCE}. Sch{\"o}lkopf et. al. ~\cite{Scholkopfetal21} discuss causal representation learning as an avenue for counterfactual explanation. 
These methods operate on supervised machine learning models and attempt to find examples close to one of the data points.

In our study, the discovery methods that identify the subgroups (e.g. anomalous subsets) are inherently 
model-agnostic search algorithms. We work with a similar research question: \textit{What features describing the individual would need to change to achieve the desired output?} ~\cite{Karimi2020ModelAgnosticCE} However, we  aim to identify the exact changes that can be made to the characteristics of a subgroup of individual to achieve a desired output (in our case, a significant change in the anomalous score). Our objective is therefore slightly different from traditional counterfactual explanation, in two ways: i) Given a subgroup that has been identified as anomalous (e.g, having a significantly higher rate of an outcome than a global average), we seek to find the least number of perturbations to the subgroup's features and feature values of that would lead to loss of anomalousness; and ii) As opposed to finding the counterfactual example, we find the feature values from the complement subset.

\section{Methods}\label{sec:methods}
\begin{figure*}[h]
	\centering
	\includegraphics[width=1.0\textwidth]{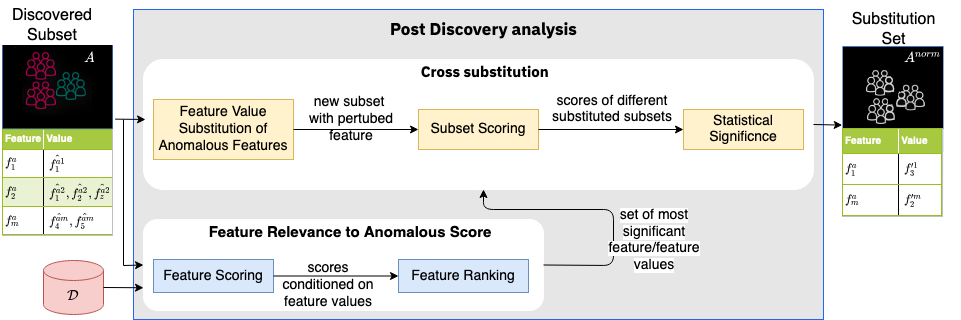}
	\caption[post-discovery Analysis Approach]{
	Pipeline for Post-Discovery Analysis of Anomalous Subsets}
	\label{fig:approach}
\end{figure*}

Figure \ref{fig:approach} illustrates our proposed pipeline for the  Post-Discovery Analysis of Anomalous Subsets. The approach takes in as input, an anomalous subset together with the original dataset and aims to identify feature substitutions that result in loss of anomalousness. It consists of two major subtasks namely: i) \textit{Feature relevance ranking}: takes an anomalous subset and identifies the most significant features and feature values contributing to anomalousness
; ii) \textit{Cross-substitution}: Identifies the possible permutations of feature values within the anomalous subsets by scoring minimally altered subsets obtained through replacing values of the anomalous features with alternatives from the nonanomalous set
. In this section, we describe these steps in detail. 

\subsection{Problem formulation}
The post-discovery analysis task assumes that there is an output from prior discovery search through a given dataset that produced a descriptive anomalous subset, and aims to identify the extent to which the feature values can be perturbed to achieve a loss of anomalousness. Formally, if we let $\mathcal{D}  = (X,Y) = \{(x_i,y_i) | i = 1,2,\cdots, N\}$ denote a dataset of length $N$, where each data-point $x_i$ is characterized by a set of $M^i$ features  $\mathcal{F}^i=[f^i_1,f^i_2,\cdots,f_{M^i}]$ such that the whole feature space set $\mathcal{F}$ in $\mathcal{D}$ is given by  $ \bigcup\limits_{i=1}^{N}\mathcal{F}_{i} $ and $y_i$ represents the outcome label,  there exists, by definition, a set $X^a  = \{(x^a_i | i = 1,2,\cdots, P\}$ s.t. $P\leq N$ and $X^a \subseteq X$, called the anomalous subset, and characterized by the anomalous feature values $\mathcal{F}^a=\{f^a_j\}_{j=1}^Z$, $Z< M$ and the complete anomalous features description is $\hat{\mathcal{F}}^a = \bigcap\limits_{z=1}^Z (\bigcup\limits_{h=1}^{H_z} \hat{f^{a}_{zh}}) $, where $\hat{f^{a}_{zh}}$ represents the $h^{th}$ value of the $z^{th}$  feature in $\mathcal{F}^a$ that contributes to the anomalousness of $X^a$, e.g., $f^a_1=$ \textit{Smoking} and $\hat{f^a_{1,1}} = $ \textit{Yes}. The tuple $A = (X^a,\mathrm{s}^a=\Gamma(X^a),\mathcal{O}^a)=S(\mathcal{D}))$ is the output of subset scanning function $S(\mathcal{D})$, where $ \Gamma(X^a) $ is a scoring function that obtains the anomalous score $s^a$, and $\mathcal{O}^a$ is a set containing the measures of effect such as the odds ratio and the p-value. The post-discovery analysis is denoted as $A^{norm} = (X^{norm},\Gamma(X^{norm}),\mathcal{O}^{norm})=g(\beta(A),\mathcal{D})$ where $g(.)$ is a cross-substitution function, that performs substitutions of the form $[\hat{f^{a}_j}\rightarrow f'_l]$, where $\hat{f^{a}_j} \in \hat{\mathcal{F}}^a$ and $f'_l \in (\hat{\mathcal{F}} \setminus \hat{\mathcal{F}}^a)$. The function $\beta$ is a feature selection routine, that identifies a set of features that most contribute to anomalousness. The output $
\hat{\mathcal{F}}^{norm} \subset (\hat{\mathcal{F}} \setminus \hat{\mathcal{F}}^a)$ is a minimalistic set of feature values that substitute in the anomalous subset to cause a loss in anomalousness. 

\subsection{Multi-Dimensional Subset Scanning(MDSS)}
\label{mdss}
To corroborate our post-discovery analysis approach to insightful analysis of anomalous subgroups, we apply the Multi-Dimensional Subset Scanning(MDSS)~\cite{Neill2013} from the anomalous pattern detection literature.  
Specifically, we use Automatic stratification (AutoStrat), a version of MDSS where the deviation between average outcomes in $\mathcal{D}$ ($\mu_D$) and each sample ($\mu_i$) is evaluated by maximizing a Bernoulli likelihood ratio scoring statistic, $\Gamma(S)$. The null hypothesis assumes that the likelihood of the outcome in each sample $x_i^r \in \mathcal{D}^r$  similar to expected ($\mu_g$), i.e., $H_0: odds(y_i)=\frac{\mu_g}{1-\mu_g}$; while the alternative hypothesis assumes a constant multiplicative increase in the outcome odds for some given subgroup, $H_1: odds(y_i)=q\frac{\mu_g}{1-\mu_g}$ where $q>1$. The scoring function, $\Gamma(S)$ for a subset, $S$, which contains $N_S$ samples, is computed as:
\[
\Gamma(S) = \max_q log(q)\sum_{i\in S} y_i - N_S * log(1-\mu_g + q\mu_g)
\]
Consequently, subsets in which the average of $y_i$ ($\mu_y$) is greater than $\mu_g$ will have higher scores.  Subset selection is iterated until convergence to a local maximum is found, and the global maximum is subsequently optimized using multiple random restarts.  Once the differentiated subset of samples $X^a$ is identified using MDSS, empirical p-value (via randomization testing) is computed to evaluate the significance of the differentiation. Moreover, subset characterization is conducted to provide interpretation of these anomalous features $\mathcal{F}^a$ and subset $X^a$.

\subsection{Quantifying Feature Relevance to Anomalousness of a Subset} 
\label{feature-selection}
The anomalous subsets produced by the subset scanning approaches described above are characterized by features and feature values contributing to the anomalousness. The intuition behind feature selection is that not all feature values contribute equally to the anomalousness of a subset, measured by the anomalous score. The feature selection, therefore, strives to identify the set of features and feature values in the anomalous subset that induce the greatest impact on the anomalous score. Such features are prime candidates for analysis as a change in their values has a higher likelihood of resulting in a significant drop in the anomalous score. Taking the Output tuple $A = (X^a,\mathrm{s}^a,\mathcal{O}^a))$ from a discovery method, and the dataset $\mathcal{D}$. The feature selection routine $\beta(.)$ proceeds by first obtaining the marginal expected values of the results conditioned on each feature value. Therefore, for each anomalous feature value, $\hat{f^{a}_j}$, we calculate its expected value as $ e_j = \mathbb E(f^{a}_j) = \frac{\sum_{i} (y_i|f^i_j=f^{ai}_j)}{|{x_i|f^i_j=f^{ai}_j}|}$.
We then calculate the deviations of $e_j$ from the mean expected value of the dataset $\bar{e} = \frac{ \sum_i y_i}{| \mathcal{D}|}$, the deviation $\delta_j=e_j - \bar{e}$. In the next step, we rank the feature values based on their calculated deviations. $rank(\hat{f^{a}_j}) < rank(\hat{f^{a}_k})$ if $\delta_j>\delta_k$ . A threshold value ($\delta_0$) is used to select a set of most deviating feature values, to form the selected features $\mathcal{F^R} = {\hat{f^{a}_j} | \delta_j > \delta_0}$.

\begin{algorithm}[H]
\footnotesize
\DontPrintSemicolon 
\KwIn{$\langle X^a,s^a,\mathcal{O}^a, \mathcal{D},\mathcal{F^R}  \rangle$; $\langle X^a \leftarrow$ anomalous subset; $\mathcal{D} \leftarrow$ dataset; $\mathcal{F^R} \leftarrow$ selected feature set}
\KwOut{($X^{norm},\mathrm{s}^{norm},\mathcal{O}^{norm}$)}
$ X^{norm} \gets X^a$\\
$\alpha \gets \alpha^a \in \mathcal{O}^a$\\
$\mathcal{F}' \gets (\mathcal{F} \setminus \mathcal{F}^a$)\\
$N \gets |\mathcal{F^R}|$\\
$ Q = enqueue(\mathcal{F^R}) $\\

\While{$s^{norm}> Threshold$ \textbf{and} $Q \neq \emptyset$}{
  $f^r =$ Q.dequeue\;
  $f'^r \gets f' \in F' | (\exists f \in F |f = f^r\cup f'r)$  \;
  \For{\textbf{each} $\hat{f^r_i} \in f^r $}{
    \For{\textbf{each} $\hat{f'^r_j} \in f'^r $}{
        $X^{norm} \gets (X^{norm} \; | \; [\hat{f^r_i} \rightarrow \hat{f'^r_j}])$\\
        $s^{norm} \gets \Gamma(X^{norm})$    // score from a Bernoulli likelihood scoring function $\Gamma(.)$ \\
        \[\Gamma(X^{norm}) = \max_q log(q)\sum_{i\in S} y_i - |X^{norm}| * log(1-\mu_g + q\mu_g)\]
        
        $\alpha \gets p\_value(s^{norm})$ \\
        $Oddr \gets oddsRatio(s^{norm},H_0)$\\
        $\mathcal{O}^{norm} = (\alpha,Oddr) $\\
    }
  }
}
\Return{($X^{norm},\mathrm{s}^{norm},\mathcal{O}^{norm}$)}\;
\caption{{\sc Cross-Substitution}}
\label{algo:cross-substitution}
\end{algorithm}

\subsection{Cross Substitution}
\label{cross-substitution}
After identifying the features and feature values that are most relevant to the anomalousness of a subset, we then perform the cross substitution process. The aim is to find minimal changes in the anomalous subset that results in a significant drop in the anomalous score. This is achieved by drawing values from the compliment set to substitute feature values in the anomalous subset. 
Assume, for example, an anomalous subset described in Table\ref{table:anomalous-subset} and its complement described in Table\ref{table:complement-subset}.

\begin{table}[!htb]
\begin{minipage}[c]{0.6\linewidth}
\centering
\caption{Anomalous Subset}
\setlength{\tabcolsep}{8pt}

\footnotesize
\begin{tabular}{cccc}
 \hline
  \textbf{Feature} & \textbf{Value(s)} & \textbf{E.g.} & \textbf{E.g.} \\
  & & Feature & Values\\
\hline
 $f^a_1$ & $\hat{f^{a1}_1}$ & \textbf{Gender} & \textit{Female} \\
 $f^a_2$ & $\hat{f^{a2}_1},\hat{f^{a2}_2}$ & \textbf{Race} & \textit{Black}, \textit{White}  \\
 $f^a_3$ & $\hat{f^{a3}_2}$ & \textbf{Smoking} & \textit{Yes}\\ 
 $f^a_4$ & $\hat{f^{a4}_2}$ & \textbf{Weight} & \textit{High}\\ 
 \hline
 
\end{tabular}
\label{table:anomalous-subset}

\end{minipage}%
\begin{minipage}[c]{0.3\linewidth}
\caption{Complement Subset}
\centering
\setlength{\tabcolsep}{8pt}
\begin{tabular}{ccc}
 \hline
  \textbf{Value(s)} & \textbf{E.g.} & \textbf{E.g.} \\
  & Feature  & Values\\
\hline
 $f'^{1}_1$ & \textbf{Gender} & \textit{Male} \\
 $f'^{2}_1$ & \textbf{Race} & \textit{Brown}  \\
 $f'^{3}_1$ & \textbf{Smoking} & \textit{No}\\ 
 $f'^{4}_1$,$f'^{4}_2$ & \textbf{Weight} & \textit{Low}, \textit{Mid}\\ 
 
 \hline
\end{tabular}

\label{table:complement-subset}
\end{minipage}
\end{table}

The possible substitutions include: \{$[\hat{f^{a}_1}\rightarrow f'^1_1]$, $[\hat{f^{a2}_1}\rightarrow f'^2_1]$, $[\hat{f^{a2}_2}\rightarrow f'^2_1]$, $[(\hat{f^{a2}_1},\hat{f^{a2}_2})\rightarrow f'^2_1]$, $[\hat{f^{a3}_2}\rightarrow f'^3_1]$, $[\hat{f^{a4}_2}\rightarrow f'^4_1]$, $[\hat{f^{a4}_2}\rightarrow f'^4_2]$ \}. A single substitution results in a new subset that is different from the anomalous subset. These perturbations are recursively carried out until we exhaust the space of possible substitutions. If this space is vastly complex, the results of the feature selection phase become invariably important.The cross-substitution therefore begins alteration in order of the feature ranking obtained from the feature selection. Each of the generated subsets are then scored with same scoring function used by the discovery method $\Gamma(.)$ (e.g. Bernoulli likelihood ratio scoring statistic). This scoring aims to generate new measures of effect $\mathcal{O}$. Statistical significance test is used to determine whether the new subset loses the anomalousness (typically a p-value statistic or odds  ratio is used, obtained i similar procedure as described in section \ref{related-work}). The stopping criterion is attained when the statistical significance of a subset surpasses a given threshold. Algorithm \ref{algo:cross-substitution} elaborates the cross-substitution analysis.

\section{Experiments and Results}
\label{experiment}
To validate our proposed methodology, we conducted a study whose objectives were twofold. First, we aimed to identify the subpopulation of osteoarthritis of the knee (OA Knee) with the most evidence of having higher-than-expected rates of major joint replacement (MJR) surgeries. Second, we aimed to identify the smallest perturbations in the characteristics of the identified anomalous subset that resulted in a significant loss of anomalousness.
 
\subsection{Dataset and Experimental Setup}
We analyzed the 2019 MarketScan Commercial Claims and Medicare data in the IBM MarketScan Research Databases (https://www.ibm.com/products/marketscan-research-databases). The database is fully deidentified, hence this study did not require Institutional Review Board approval. We defined our target cohort as newly diagnosed OA Knee patients receiving outpatient services in 2019; and our outcome cohort as outpatient OA Knee patients who underwent any MJR surgery in 2019. 
The final analytic dataset consisted of $337,078$ OA Knee patients, of whom $13,651$ ($3.9\%$) had MJR surgeries.


We extracted 5 discrete features with different cardinalities as follows: age group (18 to 34, 35 to 44, 45 to 54, 55 to 64, 65 and Older), gender (Female, Male), region (North Central, Northeast, South, Unknown, West), metropolitan statistical area (Rural, Urban), and employment status (Active Full Time, Active Part-Time or Seasonal, COBRA Continuee, Early Retiree, Long Term Disability, Medicare Eligible Retiree, Other/Unknown, Retiree (Status Unknown), Surviving Spouse/Dependendant). We defined the observed outcome as a binary indicator variable $y_i$ such that  $y_i=1$ for OA Knee patients who underwent an MJR surgery, and $y_i=0$ otherwise. We defined the expected outcome as a simple mean of the observed outcome. Accordingly, our final analysis dataset consisted of 5 features, 1 observed outcome, and one expected outcome.

\subsection{Identification and characteristics of the highest scoring (most anomalous) subset}
We used our automatic stratification approach to identify the highest-scoring subset with the most evidence of having higher rates of MJR surgeries than the global average in our dataset. 
To correct for multiple hypothesis testing and estimate the statistical significance of the identified anomalous subset, we used parametric bootstrapping to compute the empirical p-value of the subset\cite{mcfowland2013fast}. 

Using this approach, we discovered that OA Knee patients who are \textit{55 to 64 years old, reside in the West, North Central or South regions of the United States, and have an employment status as Full-time, Medicare retiree, Early Retiree, COBRA Continue, or Long Term Disability}, were most likely to undergo MJR surgeries. Among this subpopulation consisting of $135,115$ OA Knee patients, the rate of MJR surgeries was significantly higher ($6\%$ in the subpopulation compared to $3\%$ in the complement subpopulation; odds ratio 2.09, $95\%$ confidence interval, 2.02 to  2.17, p-value <0.001). 

\subsection{Post-discovery characteristics of the highest scoring (most anomalous) subset}

To perform post-discovery analysis, we first determined the relevance of features to the anomalous score. For the MJR dataset, the expected value of the output in $\mathcal{D}$ was $\bar{e} = 0.0389$. The expected value of the anomalous subset was $ e^a = 1.0 $, denoting that all records with the combination of features in the anomalous subset underwent MJR surgery. We proceeded by calculating the expected output for each feature-value in the anomalous subset. For example in the anomalous subset, one of the feature values for the feature \textit{Employment Status (EESTATU)} was \textit{"Medicare Retiree"}. We found that the expected value of the this feature value in the dataset is $e_1 = 0.057$. 

Consequently, we calculate two deviation statistics: i) A  \textit{subset deviation}: deviation of the feature value from the anomalous subset $\delta_{1\_s} = e_1 - e^a = -0.943$. ii) \textit{Global deviation}: deviation of the feature value from the expected value of the dataset $\delta_{1\_\mathcal{D}}=e_1 - e_\mathcal{D} = 0.018$. The deviation ratio of the two deviations $\delta_{r1}=\frac{\delta_{1\_s}}{\delta_{1\_\mathcal{D}}} = -52.21$ is then used to score and rank each feature value against the other feature values in the anomalous subset.

Table \ref{table:feature-import} show the results of the feature relevance and the cross-substitution steps respectively. In the feature relevance stage, we observe that the feature values with a negative deviation ratio are more relevant than a feature with positive deviation ratios. Overall the feature values with larger negative ratios are less important than those with lower negative ratio scores. For our MJR use case, the ranking of feature relevance is therefore as follows: \textit{1. Employment Medicare Retiree}, \textit{2. Early Retiree}, \textit{2. Early Retiree}.

  
\begin{table}[h]
\centering
\caption{Feature Relevance Ranking}
\resizebox{0.4\linewidth}{!}{
\begin{tabular}{ccc}\hline
  Feature & Value & D.R\\ \hline
    EESTATU & Medical Retiree & -52.21\\
    EESTATU & Early Retiree & -62.27\\
    EESTATU & L.T Disability & -78.06 \\
    EESTATU & COBRA Continue & -91.79\\
    REGION & West & -91.96\\
    AGEGRP & 55-64 & -92.20\\
    REGION & N. Cent & -92.95\\
    EESTATU & Full Time & 15293.85\\
    REGION & South & 798.75\\\hline
  \end{tabular}
\label{table:feature-import}
}
\end{table}

\begin{figure*}[h]
	\centering
	\includegraphics[width=1.0\textwidth]{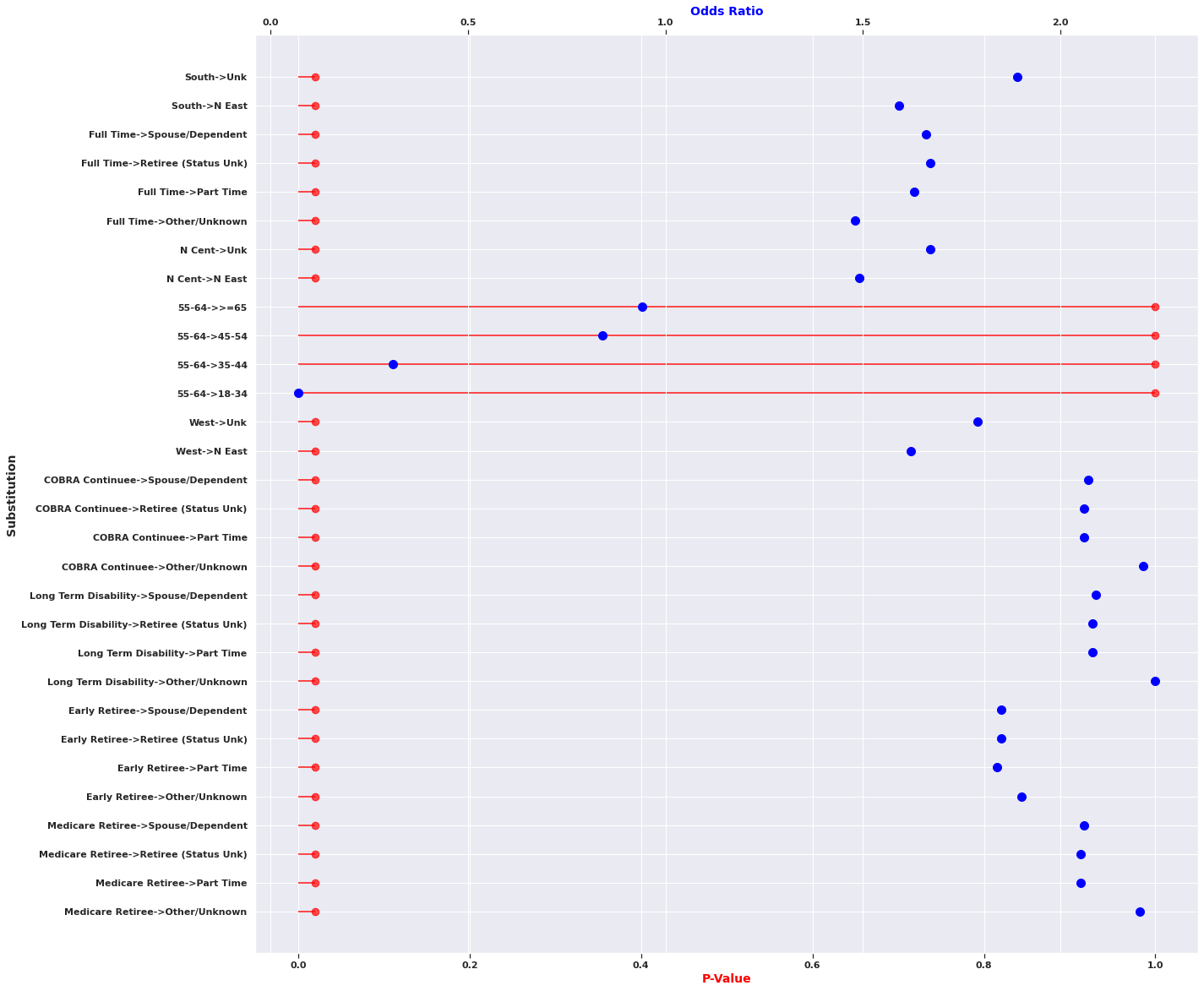}
	\caption[Single Substitution]{Plotting Cross-Substitution of Single Feature Values in the Anomalous Subset against the two Measures of Effect : P-Value and Odds Ratio}
	\label{fig:res-cross-sub}
\end{figure*}

In the Cross-Substitution stage, we perturbed the anomalous subset by changing feature values. Of note is that when we cross-substitute a feature value from the complement for an anomalous feature value, we effectively create a new subset. Consequently, we calculated a new anomalous score $s^a$, the empirical p-value of the score, and the odds ratio of the outcome in the new subset compared to its complement. For example, in the MJR dataset, the anomalous subset exhibited a score of $500.22$, empirical p-value $0.019608$, and an odds ratio of $2.09$. By substituting the employment status value \textit{"Medicare Retiree" } in the anomalous subset with  \textit{"Other Unknown" } in the complement subset, we obtain new subset with anomalous score of $465.92$, empirical p-value of $.019608$, and an odds ratio of $2.20$. These statistics are obtained using the same scoring method used in the discovery stage to allow for consistency in comparison. Table \ref{table:Substitution-scores} shows the post-discovery scores for single substitutions of the two most significant anomalous feature values. Although the p-values stay constant as seen in figure \ref{fig:res-cross-sub}, we observe a considerable degradation of the anomalous score and the Odds Ratio.
  
\begin{table}[!htb]
\centering
\caption{Observing Changes in Anomalous Scores for Substituting two Feature Values}
\setlength{\tabcolsep}{8pt}
\renewcommand{\arraystretch}{1.2}
\footnotesize
\begin{tabular}{cccccc}
 \hline
  \textbf{Feature Value} & \textbf{Substitute} & \textbf{O\_Score} & \textbf{N\_Score} & \textbf{O\_OR}  & \textbf{N\_OR} \\
\hline
 Medicare Retiree & Other/Unknown & 500.22 & 465.92 & 2.09 & 2.20 \\
 Medicare Retiree & Part Time & 500.22 & 477.93 & 2.09 & 2.05 \\
 Medicare Retiree & Retiree (Status Unk) & 500.22 & 482.87 & 2.09 & 2.05 \\
 Medicare Retiree & Spouse/Dependent & 500.22 & 483.34 & 2.09 & 2.06 \\
 Early Retiree & Other/Unknown & 500.22 & 352.94 & 2.09 & 1.86 \\
 Early Retiree & Part Time & 500.22 & 360.09 & 2.09 & 1.81 \\
 Early Retiree & Retiree (Status Unk) & 500.22 & 364.57 & 2.09 & 1.82 \\
 Early Retiree & Spouse/Dependent & 500.22 & 365.37 & 2.09 & 1.82 \\
 \hline
\end{tabular}
\label{table:Substitution-scores}
\end{table}

Figure \ref{fig:res-cross-sub} portrays the effects of making substitutions to single feature values. Two measures of effect are plotted namely: the statistical p-value of the anomalous score, and the odds ratio.  For the single substitutions, we observe that the p-value statistic remains relatively the same for most feature value perturbations. However there is a few single substitutions that cause a statistically significant change in the anomalous score namely, the substitutions on the feature: AGEGRP: \textit{55-64 -> 18-34}, \textit{55-64 -> 35-44}, \textit{55-64 -> 45-54}, \textit{55-64 -> $\geqslant65$}. From this result we can conclude thus: "if the sub-population that belong to the anomalous subgroup had been changed form the specific age group \textit{55-64} to the ages \textit{-> 18-34}, \textit{45-54}, \textit{$\geqslant65$}, there would not have been a major joint replacement surgery. Note that single value perturbations to the most relevant feature: \textit{Medicare Retiree} did not cause any statistically significant change to the anomalous score. This can be attributed to the fact that the feature employment status (EESTATU) is multi-valued in the anomalous subset, a single change in one of the feature values has little effect especially since there are 3 more feature values considered. We expect that higher combination substitutions will reveal better insights into this phenomenon. The odds ratio, on the other hand, indicates a variation of the score from the substitutions. For example the odds ratio for \textit{55-64 -> $\geqslant65$} is relatively higher than \textit{55-64 -> 18-34} meaning that the former is more likely to occur in this dataset as opposed to the latter. Similarly, the substitution: \textit{Medical Retiree -> Other Unknown}, has a very high odds ratio, attesting to the possibility of such a change in this dataset, albeit with no change in the statistical significant measure (Already explained).

\section{Conclusion}
\label{conclusion}
In this paper, we discuss a new view to interpreting population responses to disease interventions, that has the potential to improve clinical practice, As well as assist in understanding the output of subgroup analysis methods like the Autostrat. The post-discovery analysis seeks to unearth insights presented in the otherwise descriptive discovery methods outputs, by performing sufficient perturbations to the feature values. The aim thereof is to obtain the smallest changes that can lead to a loss of anomalousness.  As avenues for extension, we view our method for determining feature relevance that uses expected value weighting as a pointer characterization technique that can be extended by investigating other statistical measures related to the features in the identified subset. In addition, our current experiments on cross-substitution perform only single substitutions of the feature values. This leaves an avenue for improvement, by considering the space of all possible combinations of feature values and substitutions. Such an approach would require consideration of the cost and complexity of cross-substitution. Akin to the work in the Subset Scanning literature, we envisage the need for an approach that reduces the search in this space to Linear Time.  This work lays a foundation for the discussion concerning the design of interventions, such post-discovery analysis has great potential to inspire the better design of interventions beyond clinical Care and Public Health.

\bibliographystyle{unsrt}
{\small \bibliography{references}}


\end{document}